\documentclass[conference]{IEEEtran}
\IEEEoverridecommandlockouts

\usepackage[utf8]{inputenc} % Kullanılan encodinge göre utf8 yerine latin5 de yazılabilir.
\usepackage[T1]{fontenc}

\usepackage{algorithm,algpseudocode,caption,subcaption,setspace,colortbl}

\algnewcommand{\Inputs}[1]{%
  \State \textbf{Girdi:}
  \Statex \hspace*{\algorithmicindent}\parbox[t]{.8\linewidth}{\raggedright #1}
}
\algnewcommand{\Outputs}[1]{%
  \State \textbf{Çıktı:}
  \Statex \hspace*{\algorithmicindent}\parbox[t]{.8\linewidth}{\raggedright #1}
}

\usepackage{cite}
\ifCLASSINFOpdf
\else
\fi
\usepackage[cmex10]{amsmath}
\usepackage{multirow}
\usepackage{array}
\usepackage{subfig}

\usepackage{graphicx}
\usepackage{amsmath,algorithm,algpseudocode,epstopdf,caption,subcaption,setspace,colortbl}

\newcolumntype{M}[1]{>{\centering\arraybackslash}m{#1}}
\begin{document}

\IEEEpubid{\makebox[\columnwidth]{978-1-4673-7386-9/15/\$31.00 ©2015 IEEE\hfill}
\hspace{\columnsep}\makebox[\columnwidth]{}}

\title{Rassal Bölümlenmiş Veri Üzerinde Aşırı Öğrenme Makinesi ve Topluluk Algoritmaları ile Sınıflandırma\vspace{8pt}
Classification with Extreme Learning Machine and Ensemble Algorithms Over Randomly Partitioned Data}

\author{\IEEEauthorblockN{Ferhat Özgür Çatak}
\IEEEauthorblockA{Siber Güvenlik Enstitüsü\\
TÜBİTAK BİLGEM, Kocaeli, Türkiye\\
ozgur.catak@tubitak.gov.tr
}
}



\maketitle

\begin{ozet}
Yaşadığımız Büyük Veri çağında, makine öğrenmesi tabanlı veri madenciliği yöntemleri, yüksek boyutlu veri setlerinin analiz edilmesinde yaygın olarak kullanılmaktadır. Bu tip veri setlerinden kullanışlı tahmin modellerinin çıkarılması işlemi, yüksek karmaşıklık nedeniyle zorlayıcı bir problemdir. Veri erişiminin yüksek seviyelere ulaşmasının sağladığı fırsatla, bunların otomatik olarak sınıflandırılması önemli ve karmaşık bir görev olmaya başlamıştır. Dolayısıyla, bu bildiride güvenilir sınıflandırma tahmin model kümelerinin oluşturulması için MapReduce tabanlı dağıtık Aşırı Öğrenme Makinesi (AÖM) araştırılmıştır. Buna göre, (i) veri kümesi toplulukları oluşturulması (ii) AÖM kullanılarak zayıf sınıflandırma modellerin oluşturulması ve (iii) zayıf sınıflandırma model kümesi ile güçlü sınıflandırma modeli oluşturulmuştur. Bu eğitim yöntemi, genel kullanıma açık bilgi keşfi ve veri madenciliği veri setlerine uygulanmıştır.
%\boldmath
\end{ozet}

\begin{IEEEanahtar}
Aşırı Öğrenme Makinesi, AdaBoost, Büyük Veri, Topluluk Metodları, MapReduce
\end{IEEEanahtar}

\begin{abstract}
In this age of Big Data, machine learning based data mining methods are extensively used to inspect large scale data sets. Deriving applicable predictive modeling from these type of data sets is a challenging obstacle because of their high complexity. Opportunity with high data availability levels, automated classification of data sets has become a critical and complicated function. In this paper, the power of applying MapReduce based Distributed AdaBoosting of Extreme Learning Machine (ELM) are explored to build reliable predictive bag of classification models. Thus, (i) dataset ensembles are build; (ii) ELM algorithm is used to build weak classification models; and (iii) build a strong classification model from a set of weak classification models. This training model is applied to the publicly available knowledge discovery and data mining datasets.
%\boldmath
\end{abstract}
\vspace{-1pt}



% \fi
\IEEEpeerreviewmaketitle

\IEEEpubidadjcol
\vspace{-11pt}
\section{GİRİŞ}

Dünya genelinde bilgisayarlar, cep telefonları ve sensörler  gibi cihazlar tarafından üretilen bilgi üzerinde hem büyüklük hem de çeşit bakımından oldukça yüksek miktarda artış yaşanmaktadır. Bilgisayar teknolojisinin gelişmesiyle beraber büyük veri olarak adlandırdığımız konsept hemen her türde bilginin depolanmasına odaklanmıştır. Yüksek boyutlu veriden kullanılabilir tahmin modellerinin çıkarılması işlemi artık büyük veri kavramının içerisinde düşünülmektedir. Bu yüksek boyutlu verinin tahmin modellerinde kullanılmasının artmasıyla beraber, öğrenme algoritmalarının eğitiminin karmaşıklığıda artmaktadır. Bu nedenden dolayı, yüksek boyutlu veri setlerinin verimli bir şekilde işlenebilmesi için çeşitli topluluk metotları ve sınıflandırma algoritmalarını birleştiren makine öğrenmesi yöntemleri geliştirilmesi gerekmektedir.

Aşırı öğrenme makinesi (AÖM), Huang tarafından \cite{Huang2006489}, genelleştirilmiş tek katmanlı ileri beslemeli ağ yapısı temel alınarak geliştirilmiştir. AÖM, düşük eğitim zamanı, çok sınıflı eğitim kümelerinde yeni örnekler üzerinde yüksek genelleme özelliği ve herhangi bir eğitim parametresi içermemesi gibi avantajlarından dolayı, doküman sınıflandırma \cite{zhao2011xml}, biyoenformatik \cite{Wang2008262}, görüntü tanıma \cite{lan2013extreme} gibi bir çok farklı alanda kullanılmaktadır.

Son yıllarda araştırmacılar, tahmin modellemesi için dağıtık ve paralel çatılarla ilgili yöntemler geliştirmektedirler. Çalışmaların çok az bir kısmı MapReduce yöntemini kullanmaktadır. Bu çalışmada önerilen yöntem, yüksek boyutlu veri setlerinden tahmin modeli oluşturmak için, farklı boyutlarda rassal veri parçaları oluşturarak bunları eğitim aşamasında kullanmak, bu şekilde AÖM algoritması ve AdaBoost yöntemi ile sınıflandırma fonksiyon kümesi oluşturmaktadır. MapReduce kullanılarak, veri setinden alt veri parçaları oluşturularak eğitilen AdaBoost, topluluk yöntemleri ile birleştirilerek, tekil bir global sınıflandırma fonksiyonu ortaya çıkarılmaktadır. Çalışmanın en önemli katkıları şu şekildedir:
\vspace{-5pt}
\begin{itemize}
	\item Genelleştirilmiş MapReduce tekniği temelli AdaBoost AÖM sınıflandırma modeli ile daha hızlı ve daha iyi sınıflandırma performansına sahip model elde edilmektedir.\vspace{-5pt}
	\item Bu çalışmanın önerdiği yeni öğrenme yöntemi ile elde edilen paralel eğitim, yüksek boyutlu veri setlerinin öğrenme için kullandığı hesaplama zamanını azaltmaktadır.\vspace{-5pt}
	\item Eğitim esnasında kullanılan her bir düğüm (node) diğerinden bağımsız olmasından dolayı veri haberleşmesi azalmaktadır.
\end{itemize}

\section{ÖN BİLGİLER}
\vspace{-5pt}
Bu bölümde, çalışmada kullanılan AÖM, AdaBoost ve MapReduce hakkında bilgi verilecektir. \vspace{-8pt}
\subsection{Aşırı Öğrenme Makinesi}\label{sec:ELM}
\vspace{-5pt}
AÖM, ilk olarak tek katmanlı ileri beslemeli sinir ağı olarak geliştirilmiştir \cite{Huang2006489}. Daha sonra yapılan çalışmalarda gizli katmanın sadece nöron olmadığı genelleştirilmiş tek katmanlı ileri beslemeli ağ önerilmiştir \cite{Huang20073056}. AÖM, oluşturduğu sinir ağının giriş ağırlıkları ile gizli düğüm eğimi değerlerini rassal olarak oluşturmakta ve çıktı katmanı ağırlıkları en küçük kareler yöntemi ile hesaplamaktadır \cite{6866146}.

Bilinmeyen bir $\mathcal{X}$ dağılımından elde edilen bağımsız özdeşçe dağılmış eğitim veri kümesi $\mathcal{D}=\{(\mathbf{x}_i, y_i)\mid i=\{1,...,n\},\mathbf{x}_i \in \mathbf{R}^p,\, y_i \in \{1, 2,...,K\}\}$ olsun. Sinir ağının hedefi $f:\mathcal{X} \rightarrow \mathcal{Y}$ şeklinde fonksiyonu bulmaktır. $N$ gizli düğüme sahip tek katmanlı ileri beslemeli sinir ağı Denklem \ref{eq:slfns}'de tanımlanmıştır. 
\vspace{-5pt}
\begin{equation}
\label{eq:slfns}
f_N(\mathbf{x}) = \sum_{i=1}^{N}\beta_iG(\mathbf{a}_i,b_i,\mathbf{x}) , \, \mathbf{x} \in \mathbf{R}^n, \, \mathbf{a}_i \in \mathbf{R}^n
\end{equation}
$\mathbf{a}_i$ ve $b_i$ öğrenme parametresi, $\beta_i$ ise $i$. gizli düğümün ağırlığıdır. Genelleştirilmiş tek katman ileri besleme sinir ağı için AÖM'nin çıktı fonksiyonu Denklem \ref{eq:slfnsgen}'de gösterilmiştir.
\vspace{-5pt}
\begin{equation}
\label{eq:slfnsgen}
f_N(\mathbf{x}) = \sum_{i=1}^{N}\beta_iG(\mathbf{a}_i,b_i,\mathbf{x}) = \mathbf{\beta} \times h(\mathbf{x})
\end{equation}
İkili sınıflandırma uygulamaları için AÖM karar fonksiyonu ise Denklem \ref{eq:binaryelm}'de gösterilmiştir.
\vspace{-5pt}
\begin{equation}
\label{eq:binaryelm}
f_N(\mathbf{x}) = sign\left( \sum_{i=1}^{N}\beta_iG(\mathbf{a}_i,b_i,\mathbf{x}) \right) = sign\left(\mathbf{\beta} \times h(\mathbf{x}) \right)
\end{equation}
Denklem \ref{eq:slfnsgen} diğer bir form olarak Denklem \ref{eq:elm}'de gösterilmiştir.
\vspace{-5pt}
\begin{equation}
\label{eq:elm}
H\beta=T
\end{equation}
$H$ ve $T$ sırasıyla gizli katman matrisi ve çıktı matrisidir. Gizli katman matrisi Denklem \ref{eq:H}'de gösterilmiştir.
\vspace{-5pt}
\begin{equation}
\label{eq:H}
H(\tilde{a},\tilde{b},\tilde{x})= \begin{bmatrix} G(a_1,b_1,x_1) & \cdots & G(a_l,b_l,x_1) \\ \vdots & \ddots & \vdots \\ G(a_1,b_1,x_N) & \cdots & G(a_l,b_l,x_N) \end{bmatrix}_{N \times L}
\end{equation}
burada $\tilde{a}=a_1,...,a_L$, $\tilde{b}=b_1,...,b_L$, $\tilde{x}=x_1,...,x_N$ şeklindedir. Çıktı matrisi Denklem \ref{eq:elmoutput}'de gösterilmiştir.
\vspace{-5pt}
\begin{equation}
\label{eq:elmoutput}
T= \begin{bmatrix} t_1 \hdots t_N \end{bmatrix}
\end{equation}
\vspace{-5pt}
\subsection{AdaBoost}\label{sec:AdaBoost}
\vspace{-5pt}
AdaBoost, nitelik matrisi $X$ ve çıktı sınıfları, $y \in \{+1, -1\}$, kullanarak, zayıf öğrenme modelleri, $h_t(\mathbf{x})$, birleştirerek $H(\mathbf{x})$ şeklinde güçlü sınıflandırma modeli oluşturmaya çalışmaktadır \cite{LandesaVazquez2013101}. Denklem \ref{eq:adaboost}'de güçlü sınıflandırma modeli gösterilmektedir.
\vspace{-5pt}
\begin{equation}
\label{eq:adaboost}
H(\mathbf{x}) = sign(f(\mathbf{x}))=sign\left(\sum_{t=1}^{T}\alpha_t h_t(\mathbf{x}) \right)
\end{equation}

\subsection{MapReduce}\label{sec:MapReduce}
MapReduce yöntemi yüksek boyutlu veri setlerinin işlenmesine olanak sağlayan, ayrıca Google tarafından da oldukça sık kullanılan bir programlama modelidir \cite{Dean:2008:MSD:1327452.1327492}. Kullanıcılar tarafından tanımlanan $Map$ ve $Reduce$ fonksiyonları ve bu fonksiyonlara girdi değeri olarak verilen anahtar/değer dizileri (Key/Value pairs) kullanılmaktadır. MapReduce yönteminde kullanılan anahtar değer veri modeli genellikle ilişkisel veri modelleri ile tasarlanamayacak veri setlerine uygulanmaktadır. Örnek olarak bir web sayfasının adresi anahtar değerine yazılırken bu sayfanın HTML içeriği ise değer alanına yazılmaktadır. Grafik tabanlı veri modellerinde ise anahtar alanı düğüm anahtar (id) bilgisini içerirken değer ise liste olarak kendisine komşu olan düğümlerin anahtar bilgilerini içerebilir.

$Map$ fonksiyonu paralel olarak girdi veri setinde bulunan her ikiliye uygulanmaktadır. Fonksiyon bir veri alanında bulunan veri çiftlerini alarak bunları farklı bir alana veri çift listesi olarak vermektedir.
\begin{equation}
Map(a_1, d_1) \rightarrow liste(a_2,d_2)
\end{equation}
$Reduce$ fonksiyonu ise yine paralel olarak $Map$ fonksiyonu tarafından ilişkilendirilmiş anahtar değer yapısına uygulayarak yeni değerler listesi oluşturmaktadır. 
\begin{equation}
Reduce(a_2, liste(d_2)) \rightarrow liste(a_3,d_3)
\end{equation}
MapReduce çatısının anahtar/değer şeklindeki çiftlerden oluşan listeyi değerler listesi şekline çevirmektedir.
\vspace{-15pt}
\section{SİSTEM MODELİ}
Bu bölümde, MapReduce temelli AdaBoost AÖM algoritmasının detayları verilecektir. Temel fikir Bölüm \ref{sec:BasicIdea} kısmında anlatılacaktır. Sistemin gerçekleştirimi ise Bölüm \ref{sec:Implementation} kısmında tanımlanacaktır.
\vspace{-15pt}
\subsection{Temel Fikir}\label{sec:BasicIdea}
\vspace{-5pt}
AdaBoost temelli AÖM sınıflandırma algoritmasının hesaplanması aşamasının dağıtık ve paralel hale getirilmesi bu çalışmanın esas görevidir. Önerilen yöntemin temel fikri, sınıflandırma topluluk fonksiyonlarının rassal veri parçaları $(X_m,Y_m)$ kullanılarak paralel olarak hesaplanmasıdır.

Tablo \ref{tbl:notation}'de, bildirinin anlaşılmasında kolaylık olması için çalışmada kullanılan değişken ve notasyonların özeti verilmiştir.
\begin{table}[!t] %[ph]
	\footnotesize
	\caption{Sık kullanılan değişkenler ve notasyonlar.}
	\label{tbl:notation}
	\begin{center}
		\begin{tabular}{|M{1cm}|M{2.5cm}||M{1cm}|M{2.5cm}|}
		\hline \textbf{Notasyon} & \textbf{Açıklama} & \textbf{Notasyon} & \textbf{Açıklama} \\ 
		\hline
		$M$ & Veri parça bölümleme uzunluğu & $T$ & AdaBoost $T$ boyutu \\ 
		\hline
		$h$ & Sınıflandırma fonksiyonu & $nh$ & AÖM'de kullanılan gizli düğüm sayısı \\
		\hline
		$X_m$ & Veri seti, $\mathcal{D}$, girdi değerlerinin $m$ veri parçası & Doğ & Sınıflandırma hipotezinin doğruluğu \\
		\hline
		$Y_m$ & Veri seti, $\mathcal{D}$, çıktı değerlerinin $m$ veri parçası  & H. & Hassasiyet \\
		\hline
		$\epsilon$ & Hata oranı & G.C. & Geri Çağırım \\
		\hline 
		\end{tabular} 
	\end{center}
\end{table}

\subsection{Modelin Gerçekleştirimi}\label{sec:Implementation}
\vspace{-5pt}
MapReduce temelli AdaBoost AÖM algoritmasının sözde kodu Algoritma \ref{alg:map} ve Algoritma \ref{alg:reduce}'de gösterilmiştir. Önerilen öğrenme modelinin $Map$ metodu, bölümleme boyutu, $M$, aralığına kadar tam sayı olacak şekilde rassal değerin eğitim veri kümesinin her bir satırına atanması şeklindedir. $Map$'in girdi değeri olan $\mathbf{x}$, eğitim veri kümesi $\mathcal{D}$'nin bir satırıdır. $Map$ metodu girdi matrisini satır olarak bölümlemekte ve $<rassalBolumId,\mathbf{x}>$ anahtar/değer ikililerini oluşturmaktadır. $rassalBolumId$, veri parçasının tanımlayıcısı olarak atanmakta ve anahtar olarak $Reduce$ aşamasına transfer edilmektedir.
\begin{algorithm}[h]
	\caption{AdaBoostAÖM::Map}\label{alg:map}
	\begin{algorithmic}[1]
		\Inputs{$(\mathbf{x},y) \in \mathcal{D}$, $M$}
		\State $k \gets rand(0,M)$
		\State $Output(k,(\mathbf{x},y))$
	\end{algorithmic}
\end{algorithm}
$Reduce$ aşamasının sözde kodu Algoritma \ref{alg:reduce}'de gösterilmiştir. $Reduce$ aşaması, Algoritma \ref{alg:reduce} sözde kodunun 3 -- 8. satırları arasında bulunan döngüde gerçekleştirilmiştir. Her bir veri parçası, $(\mathbf{X}_k,\mathbf{y}_k)$, AdaBoost topluluk yöntemi temelli AÖM ile eğitilmektedir. Böylece her bir $Reduce$ işlemi ayrı bir sınıflandırma modeli ortaya çıkarmaktadır. $Reduce$ işleminde anahtar, $k$, $Map$ aşamasında rassal olarak atanan $rassalBolumId$, girdi olarak kullanılmaktadır.

\begin{algorithm}[h]
	\caption{AdaBoostAÖM::Reduce}\label{alg:reduce}
	\begin{algorithmic}[1]
		\Inputs{Anahtar $k$, Deger $V$, $T$}
		\State $(\mathbf{X}_n,\mathbf{y}_n) \gets V$
		\For{$t=1..T$}
			\State $h_t \gets AOM(\mathbf{X}_n,\mathbf{y}_n)$
			\State $\mathbf{y}_{pred},\epsilon_t \gets h_t(\mathbf{X}_n)$
			\State $\alpha_t \gets  \frac{1}{2}\ln{\frac{1-\epsilon_t}{\epsilon_t}}$
			\State $\mathcal{D}_{t+1} = \frac{\mathcal{D}_t \times exp(-\alpha_ty_ih_t(x_i))}{Z_t}$
		\EndFor
		\Outputs{$h_m = sign\left(\sum_{t=1}^{T}{\alpha_t h_t(\mathbf{x}) }\right) $}.
	\end{algorithmic}
\end{algorithm}

\vspace{-10pt}
\section{BENZETİM SONUÇLARI}\label{sec:experiments}
Bu bölümde, internet ortamında açık olarak erişilebilen gerçek veri setleri kullanılarak, önerilen modelin sınıflandırma performansı farklı ölçüm yöntemleri ile sınanmıştır. Gerçekleştirim aşamasında 64 bit Python 2.7 yazılım dili ve MrJob kütüphanesi kullanılmıştır.

Bölüm \ref{sec:expsetup}'da, deneysel ortamda kullanılan veri setleri ve AÖM'nin parametreleri açıklanmaktadır. Standart AÖM'nin her bir veri setinde sınıflandırma performanslarının saklı düğümlere göre değişimi Bölüm \ref{sec:conv_elm}'de gösterilmektedir. Bölüm \ref{sec:eval}'de, önerilen öğrenme modelinin deneysel sonuçları tablo ve grafik olarak gösterilmektedir.
\vspace{-10pt}
\subsection{Deneysel Kurulum}\label{sec:expsetup}
\vspace{-3pt}
Bu bölümde, önerilen yöntem, Pendigit, Letter, Statlog ve Page-blocks şeklinde dört farklı veri seti kullanılarak sınıflandırma modeli oluşturulmuş, bu şekilde yöntemin etkinliği ve verimliliği ölçümlenmeye çalışılmıştır. Kullanılan açık dört farklı veri kümesi Tablo \ref{tbl:dslist}'de gösterilmektedir. Kullanılan bütün veri setleri, ikiden fazla etikete sahip, çok sınıflıdır. 

\begin{table}[!t] %[ph]
	\caption{Kullanılan veri setlerinin bilgileri.}
	\label{tbl:dslist}
	\begin{center}
		\begin{tabular}{|c|r|r|c|c|}
		\hline
		\textbf{Veri seti} & \textbf{\# Eğitim} & \textbf{\# Test} & \textbf{\# Sınıf} & \textbf{\# Öz nitelik} \\ 
		\hline Pendigit & 7495 & 3498 & 10 & 64 \\ 
		Skin & 220543 & 24507 & 2 & 4 \\ 
		Statlog & 43500 & 25000 & 10 & 7 \\ 
		Page-blocks & 4500 & 973 & 5 & 10 \\
		%\hline Waveform & 4400 & 600 & 3 & 21 \\
		\hline 
		\end{tabular} 
	\end{center}
\end{table}
\vspace{-10pt}
\subsection{Veri setlerinin Standart AÖM ile sonuçları}\label{sec:conv_elm}
Tablo \ref{tbl:convelm}'de çalışmanın deneysel kısmında kullanılan veri setlerinin AÖM sonuçları paylaşılmıştır. $nh$ değeri, 1 -- 500 arasında değişmektedir. Performans ölçümleri için doğruluk, hassasiyet, geri çekilme ve $F_1$ değerleri kullanılmıştır. 
\begin{table}[ph]
	\caption{Veri setlerinin standart AÖM sonuçları.}
	\label{tbl:convelm}
	\begin{tabular}[t]{|c|r|r|r|r|r|}
	\hline
	Veri S. & $nh$. & Doğ. & H. & G.C. & $F_1$ \\
	\hline
	Pendigit & 149 & 0,8404 & 0,8393 & 0,8416 & 0,8407 \\
	Skin & 98 & 0,9754 & 0,9956 & 0,9583 & 0,9894 \\
	Statlog & 249 & 0,8871 & 0,8556 & 0,9237 & 0,9757 \\
	Page Blocks & 498 & 0,9873 & 0,9794 & 0,9988 & 0,9977 \\
	\hline
	\end{tabular}
\end{table}

%\begin{figure*}
%	\begin{subfigure}[b]{0.23\textwidth}
%		\includegraphics[width=1\linewidth]{fig/statlog_conv}
%		\caption{Statlog data set.}
%		\label{fig:statlog_conv}
%	\end{subfigure}
%	\hspace{1pt}
%	\begin{subfigure}[b]{0.23\textwidth}
%		\includegraphics[width=1\linewidth]{skin_conv}
%		\caption{Skin data set.}
%		\label{fig:skin_conv}
%	\end{subfigure}
%	\hspace{1pt}
%	\begin{subfigure}[b]{0.23\textwidth}
%	\includegraphics[width=1\linewidth]{pendigit_conv}
%	\caption{Pen digit data set.}
%	\label{fig:pendigit_conv}
%	\end{subfigure}
%%	\hspace{1pt}
%%	\begin{subfigure}[b]{0.15\textwidth}
%%	\includegraphics[width=1\linewidth]{waveform_conv}
%%	\caption{Waveform data set.}
%%	\label{fig:waveform_conv}
%%	\end{subfigure}
%	\hspace{1pt}
%	\begin{subfigure}[b]{0.23\textwidth}
%	\includegraphics[width=1\linewidth]{page-blocks_conv}
%	\caption{Page blocks data set.}
%	\label{fig:page-blocks_conv}
%	\end{subfigure}
%	
%	\caption{Number of hidden nodes in ELM versus classifier precision.}
%	\label{fig:convelm}
%\end{figure*}
\vspace{-20pt}
\subsection{Sonuçlar}\label{sec:eval}
Bu çalışma kapsamında kullanılan veri setlerinin sınıf dağılımları dengesiz olmasından dolayı optimal sınıflandırıcı hipotezin bulunmasında sadece geleneksel doğruluk tabanlı performans ölçümü yeterli değildir. Bu nedenle sınıflandırıcı hipotez ölçümünde ortalama doğruluk, ortalama hassasiyet, ortalama geri çekim \cite{Turpin:2006:UPV:1148170.1148176}, $F_1$ ölçümü şeklinde dört farklı yöntem kullanılmıştır. Kullanılan yöntemler bilgi çıkarımı alanında en çok kullanılan yöntemlerdir \cite{Makhoul99performancemeasures}.

Hassasiyet, elde edilen ilgili örneklerin toplam elde edilen örneklere oranıdır. Hassasiyet Denklem \ref{eqn:prec}'da gösterilmektedir.
% Precision is defined as the fraction of retrieved samples that are relevant. Precision is shown in Eq. \ref{eqn:prec}. 
\begin{equation}
	\label{eqn:prec}
	Hassasiyet = \frac{Dogru}{Dogru + Hata}
\end{equation}
Geri çekilme, elde edilen ilgili örneklerin toplam ilgili örneklere oranıdır. Geri çekilme Denklem \ref{eqn:recall}'de gösterilmektedir.
% Recall is defined as the fraction of relevant samples that is retrieved. Recall is shown in Eq. \ref{eqn:recall}.
\begin{equation}
	\label{eqn:recall}
	Geri \, Cekilme = \frac{Dogru}{Dogru + Kayip}
\end{equation}
Önerilen modelde, her bir sınıf için farklı olarak hassasiyet ve geri çekilme değerleri hesaplanıp  toplam sınıf sayısına bölünerek, elde edilen sınıflandırma hipotezinin ortalama ölçüm değerleri hesaplanmaktadır. Ortalama hassasiyet ve geri çekilme Denklem \ref{eqn:avgprec} ve Denklem \ref{eqn:avgrecall}'de gösterilmektedir.

\begin{equation}
	\label{eqn:avgprec}
	Hassasiyet_{ort} = \frac{1}{n_{sinif}}\sum_{i=0}^{n_{sinif}-1}{Hassasiyet_i}
\end{equation}
\begin{equation}
	\label{eqn:avgrecall}
	Geri\,Cekilme_{ort} = \frac{1}{n_{sinif}}\sum_{i=0}^{n_{sinif}-1}{Geri Cekilme_i}
\end{equation}
$F_1$ ölçümü, hassasiyet ve geri çekilmenin harmonik ortalamasıdır. Değerlendirme modeli, Denklem \ref{eqn:fmeasure}'de gösterilen $F_1$ ölçümünün çok sınıflı halini kullanmaktadır.
\begin{equation}
	\label{eqn:fmeasure}
	F_1 = 2 \times \frac{Hassasiyet_{ort} \times Geri Cekilme_{ort}}{Hassasiyet_{ort} + Geri Cekilme_{ort}}
\end{equation}

\begin{figure}[h]
	\begin{subfigure}[b]{0.15\textwidth}
		\includegraphics[width=1\linewidth]{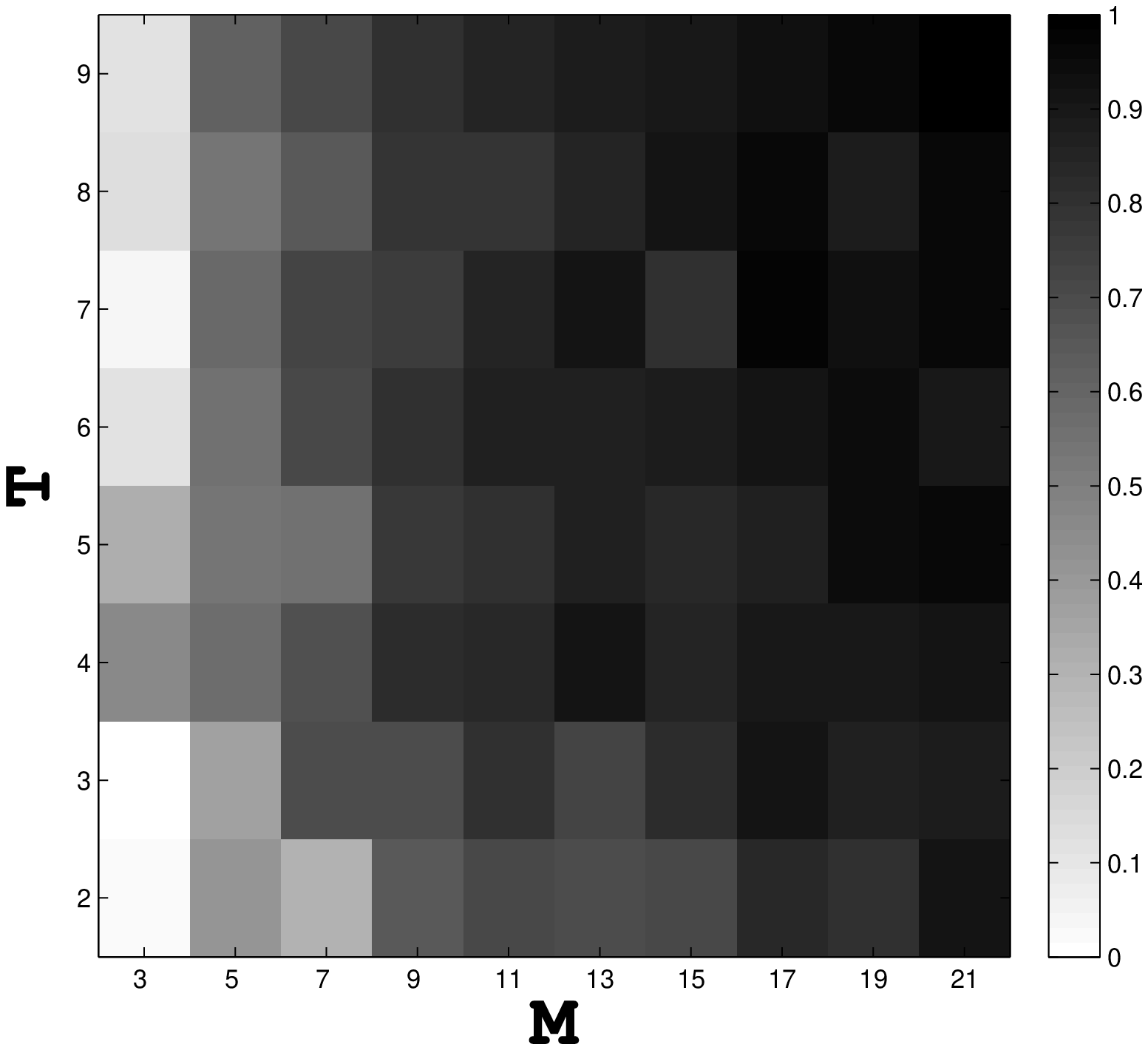}
		\caption{Bölümleme ve adaboost $T$ boyu}
		\label{fig:statlog_map_T}
	\end{subfigure}
	\hspace{1pt}
	\begin{subfigure}[b]{0.15\textwidth}
		\includegraphics[width=1\linewidth]{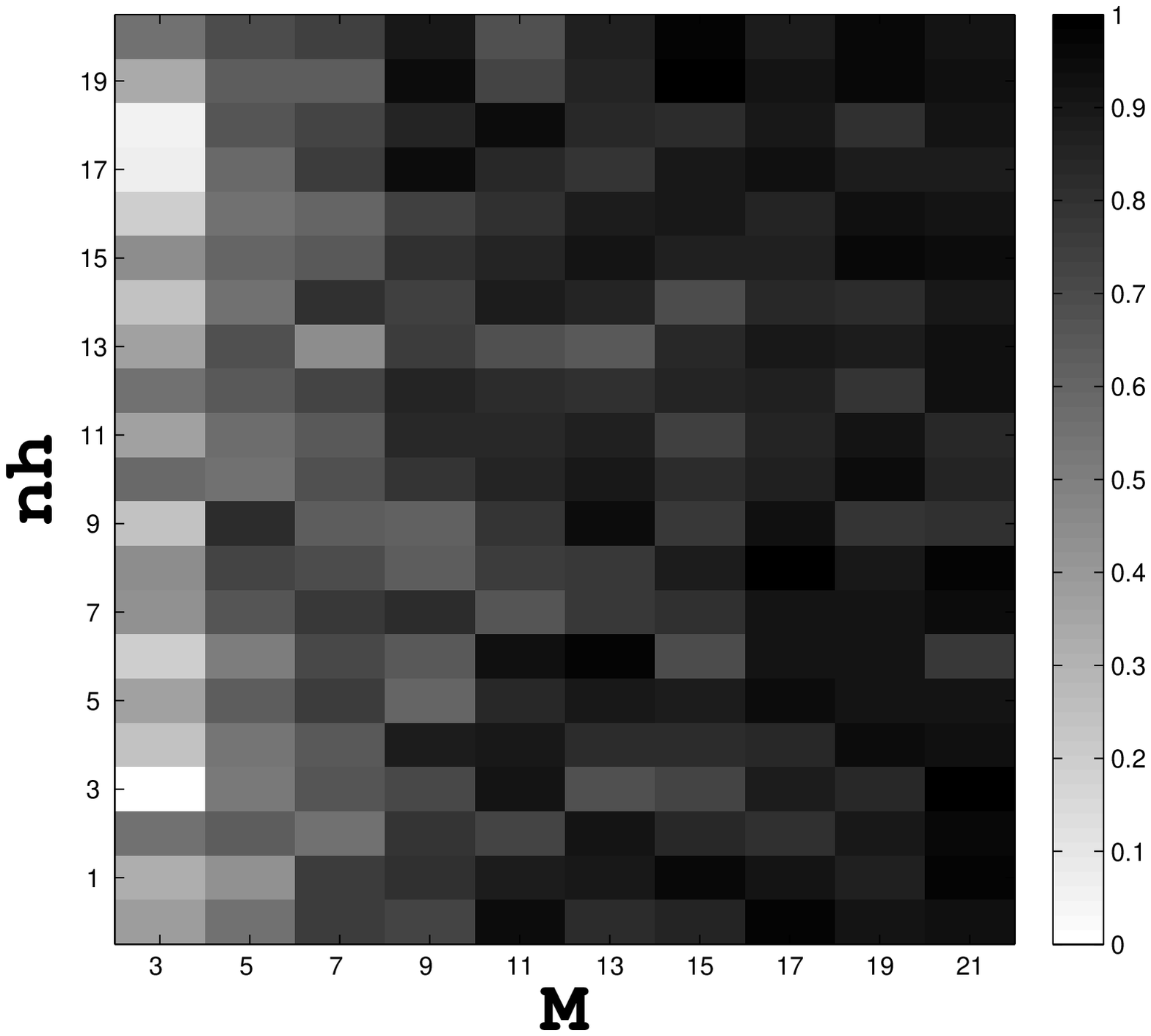}
		\caption{Bölümleme ve gizli düğüm sayısı.}
		\label{fig:statlog_map_nh}
	\end{subfigure}
	\hspace{1pt}
	\begin{subfigure}[b]{0.15\textwidth}
	\includegraphics[width=1\linewidth]{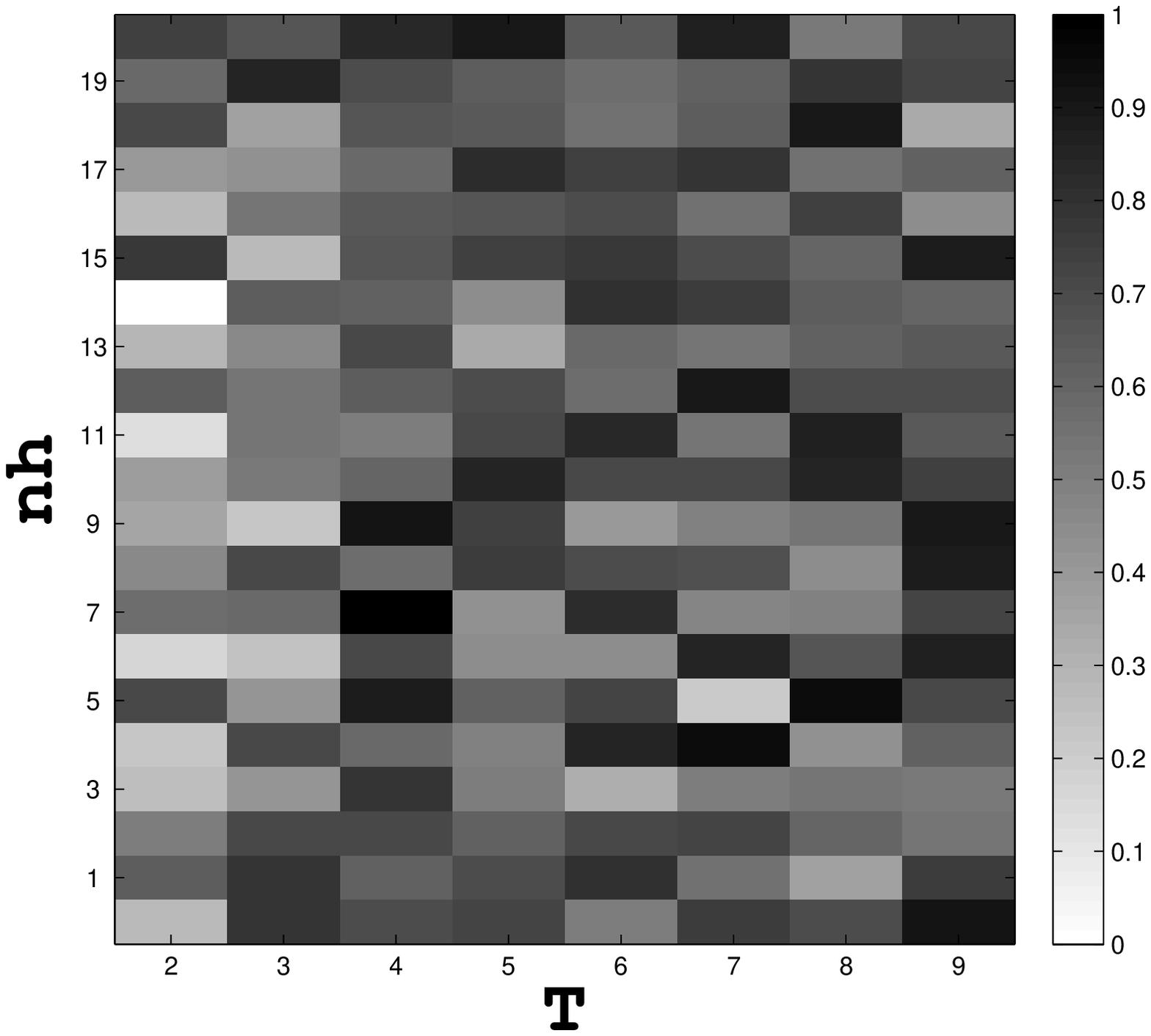}
	\caption{AdaBoost $T$ boyu ve gizli düğüm sayısı.}
	\label{fig:statlog_T_nh}
	\end{subfigure}
	\caption{Statlog veri kümesi ısı haritası.}
	\label{fig:statlogres}
\end{figure}

\begin{figure}[h]
	\begin{subfigure}[b]{0.15\textwidth}
		\includegraphics[width=1\linewidth]{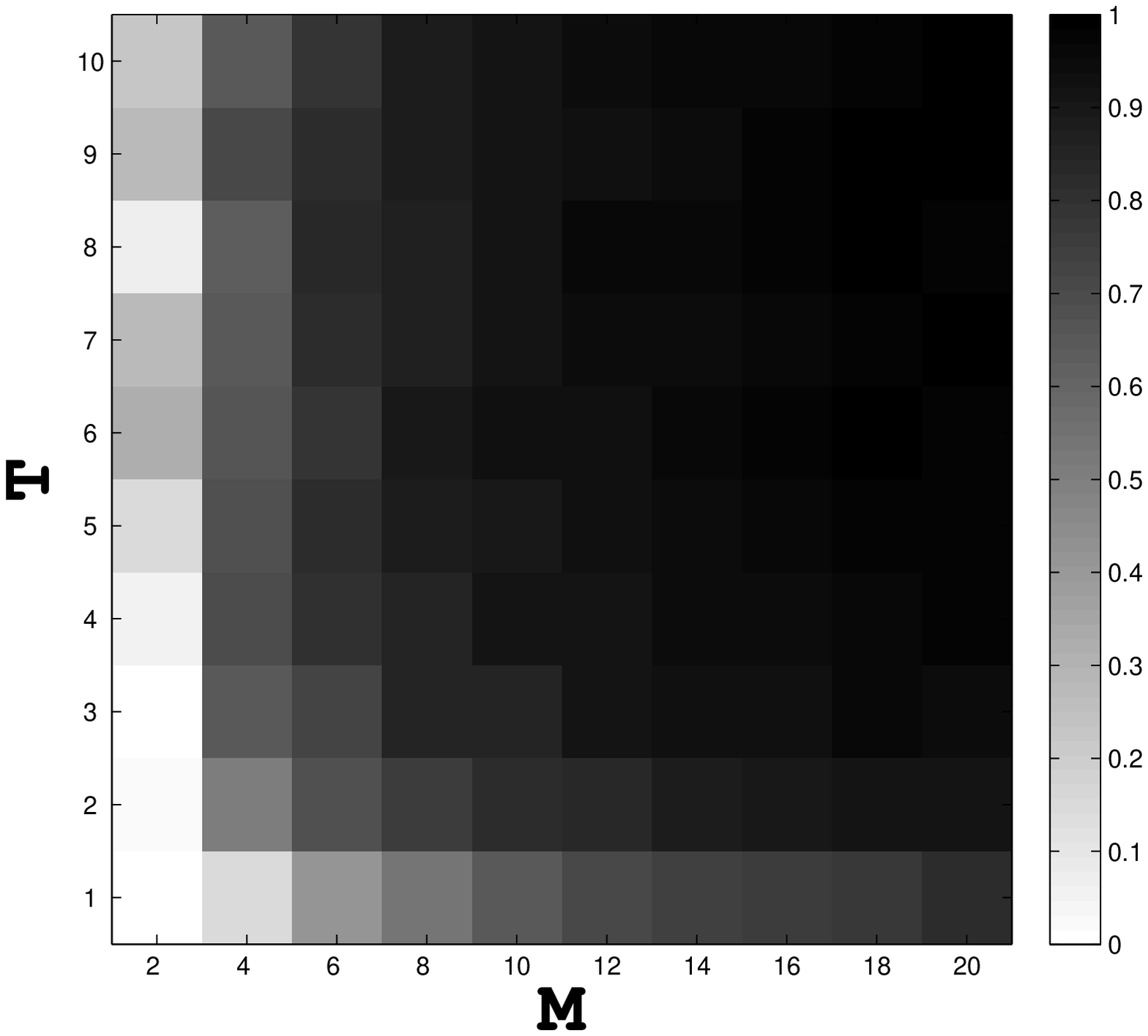}
		\caption{Bölümleme ve adaboost $T$ boyu}
		\label{fig:pendigit_map_T}
	\end{subfigure}
	\hspace{1pt}
	\begin{subfigure}[b]{0.15\textwidth}
		\includegraphics[width=1\linewidth]{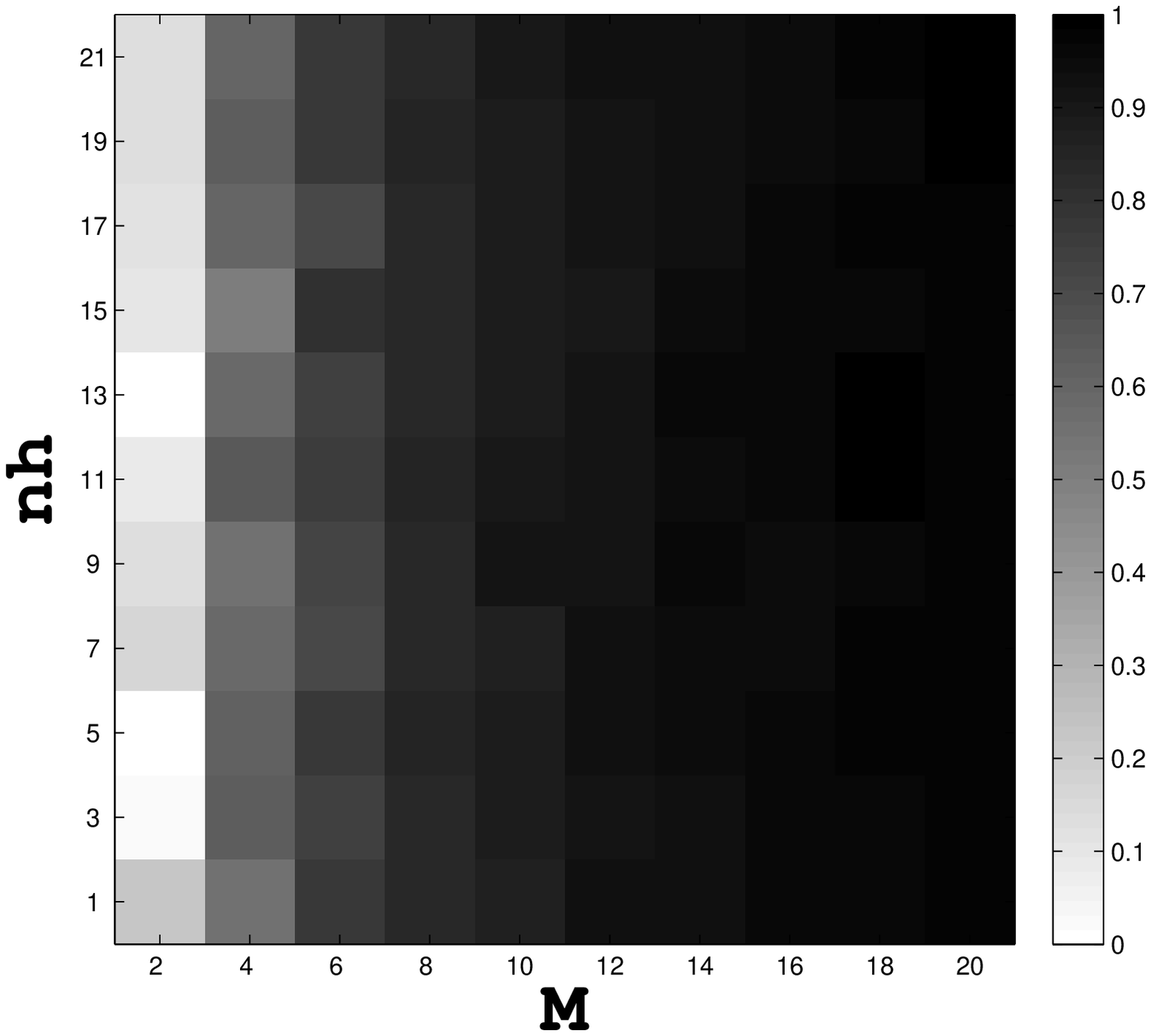}
		\caption{Bölümleme ve gizli düğüm sayısı.}
		\label{fig:pendigit_map_nh}
	\end{subfigure}
	\hspace{1pt}
	\begin{subfigure}[b]{0.15\textwidth}
	\includegraphics[width=1\linewidth]{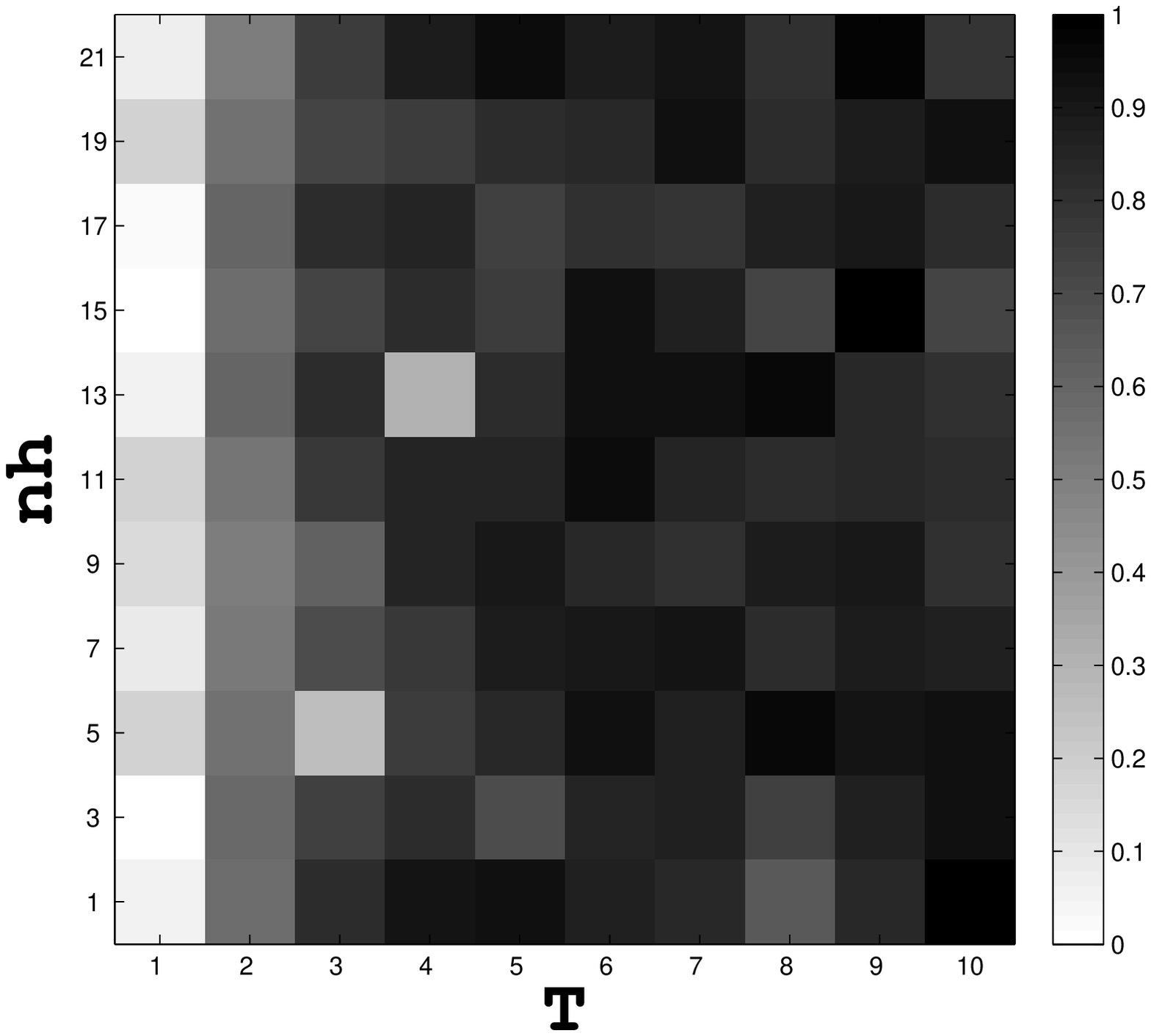}
	\caption{AdaBoost $T$ boyu ve gizli düğüm sayısı.}
	\label{fig:pendigit_T_nh}
	\end{subfigure}
	\caption{Pendigit veri kümesi ısı haritası.}
	\label{fig:pendigitres}
\end{figure}

\begin{figure}[h]
	\begin{subfigure}[b]{0.15\textwidth}
		\includegraphics[width=1\linewidth]{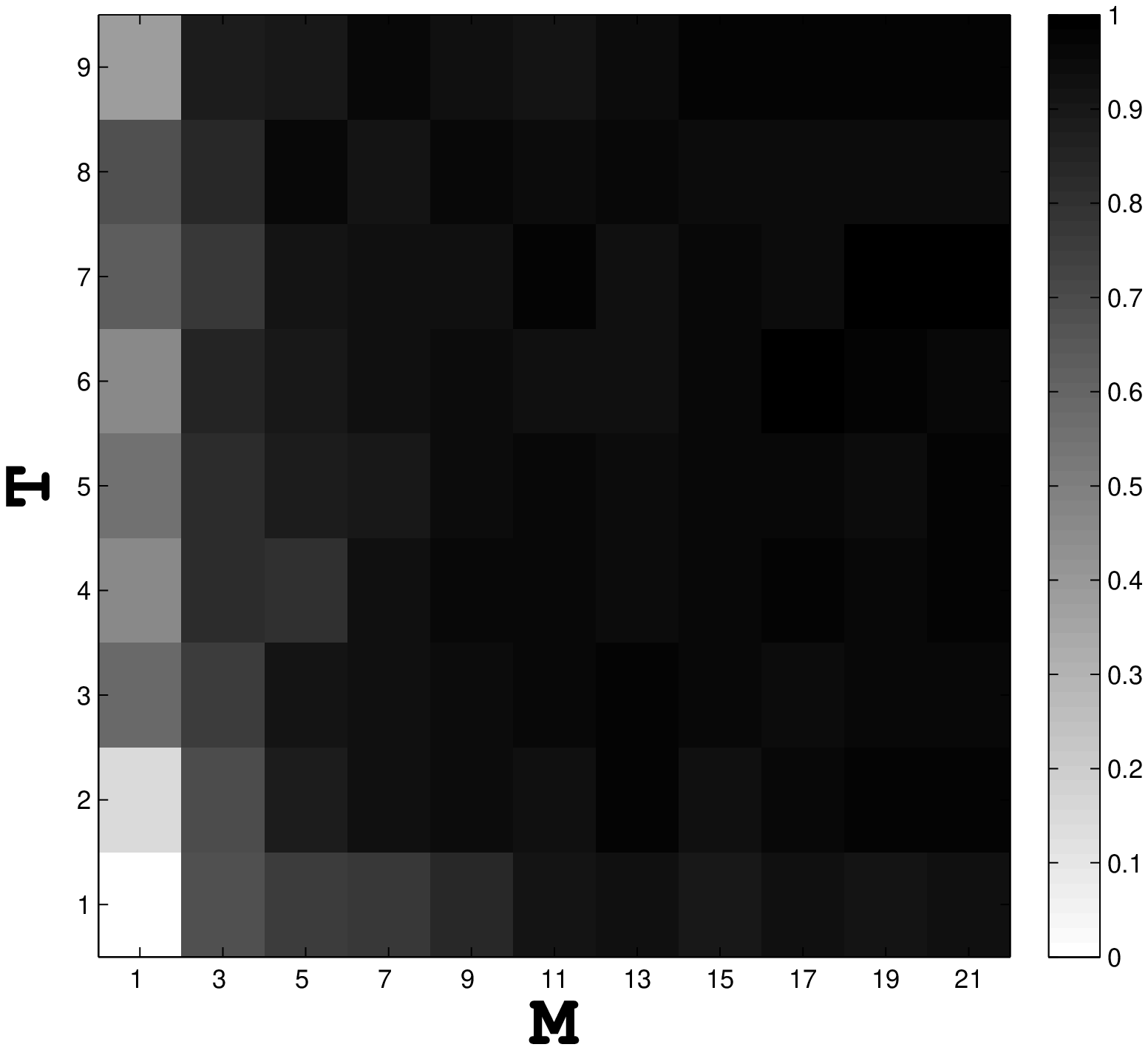}
		\caption{Bölümleme ve adaboost $T$ boyu}
		\label{fig:skin_map_T}
	\end{subfigure}
	\hspace{1pt}
	\begin{subfigure}[b]{0.15\textwidth}
		\includegraphics[width=1\linewidth]{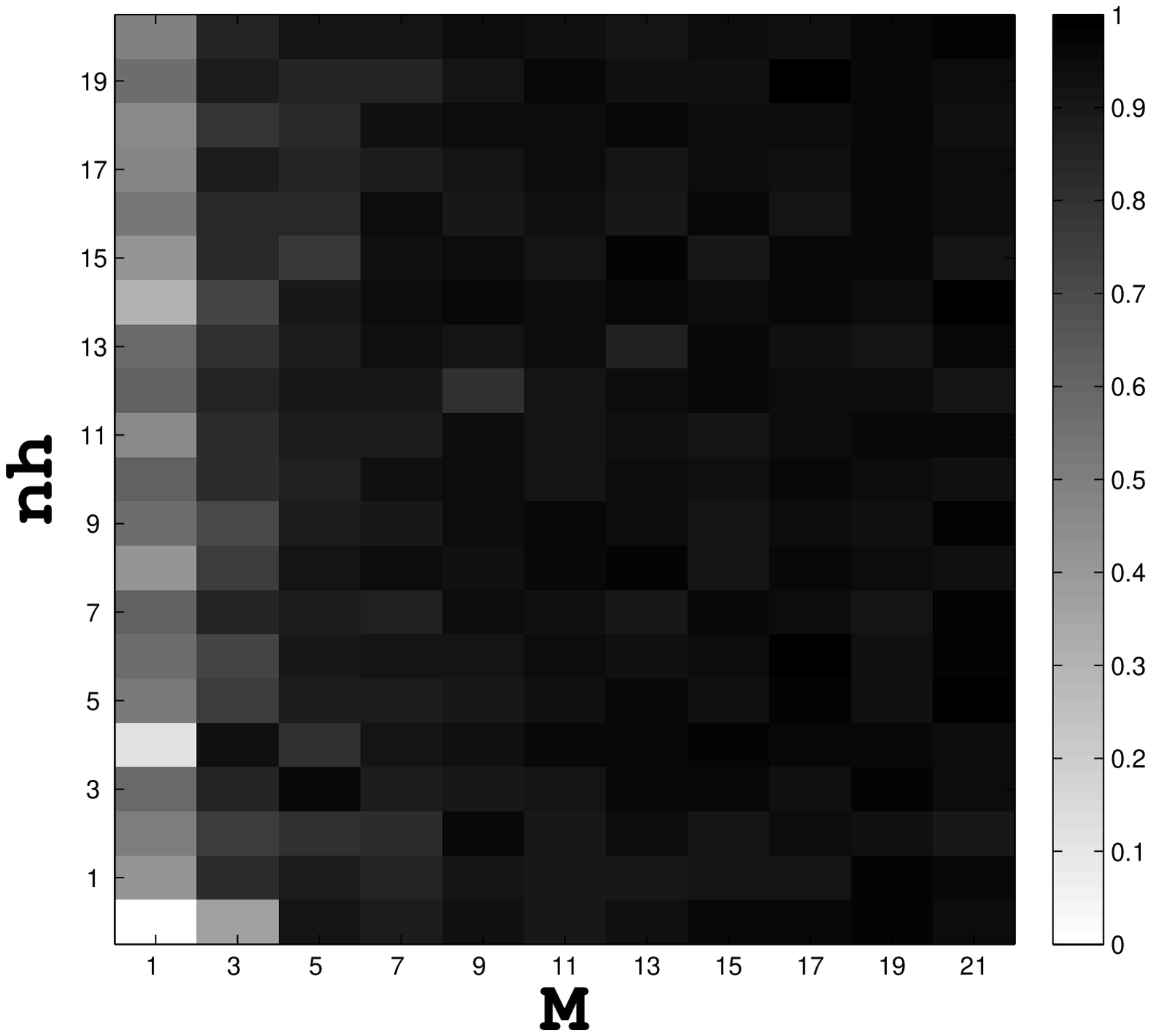}
		\caption{Bölümleme ve gizli düğüm sayısı.}
		\label{fig:skin_map_nh}
	\end{subfigure}
	\hspace{1pt}
	\begin{subfigure}[b]{0.15\textwidth}
	\includegraphics[width=1\linewidth]{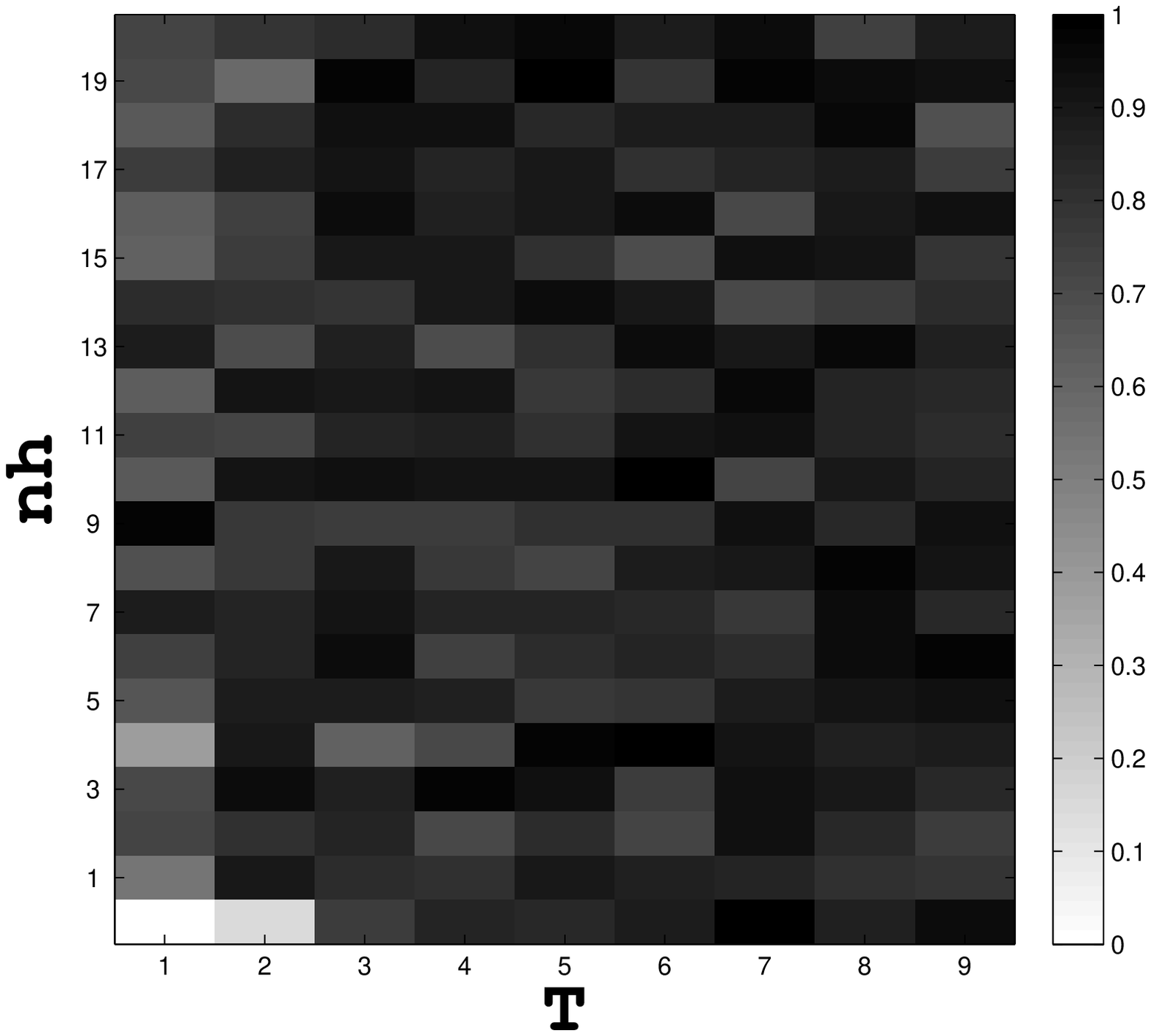}
	\caption{AdaBoost $T$ boyu ve gizli düğüm sayısı.}
	\label{fig:skin_T_nh}
	\end{subfigure}
	\caption{Skin veri kümesi ısı haritası.}
	\label{fig:skinres}
\end{figure}

\begin{figure}[h]
	\begin{subfigure}[b]{0.15\textwidth}
		\includegraphics[width=1\linewidth]{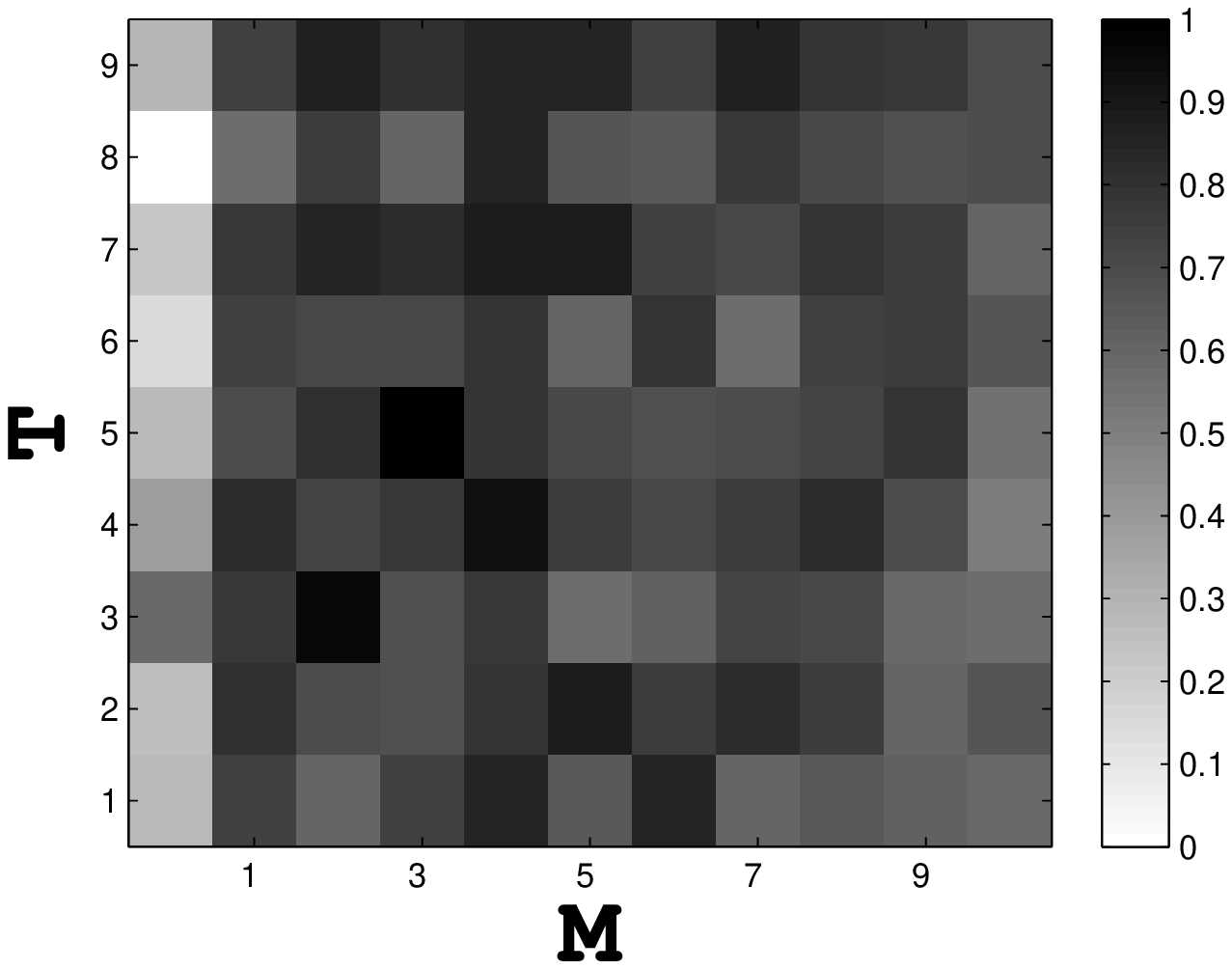}
		\caption{Bölümleme ve adaboost $T$ boyu}
		\label{fig:pageblocks_map_T}
	\end{subfigure}
	\hspace{1pt}
	\begin{subfigure}[b]{0.15\textwidth}
		\includegraphics[width=1\linewidth]{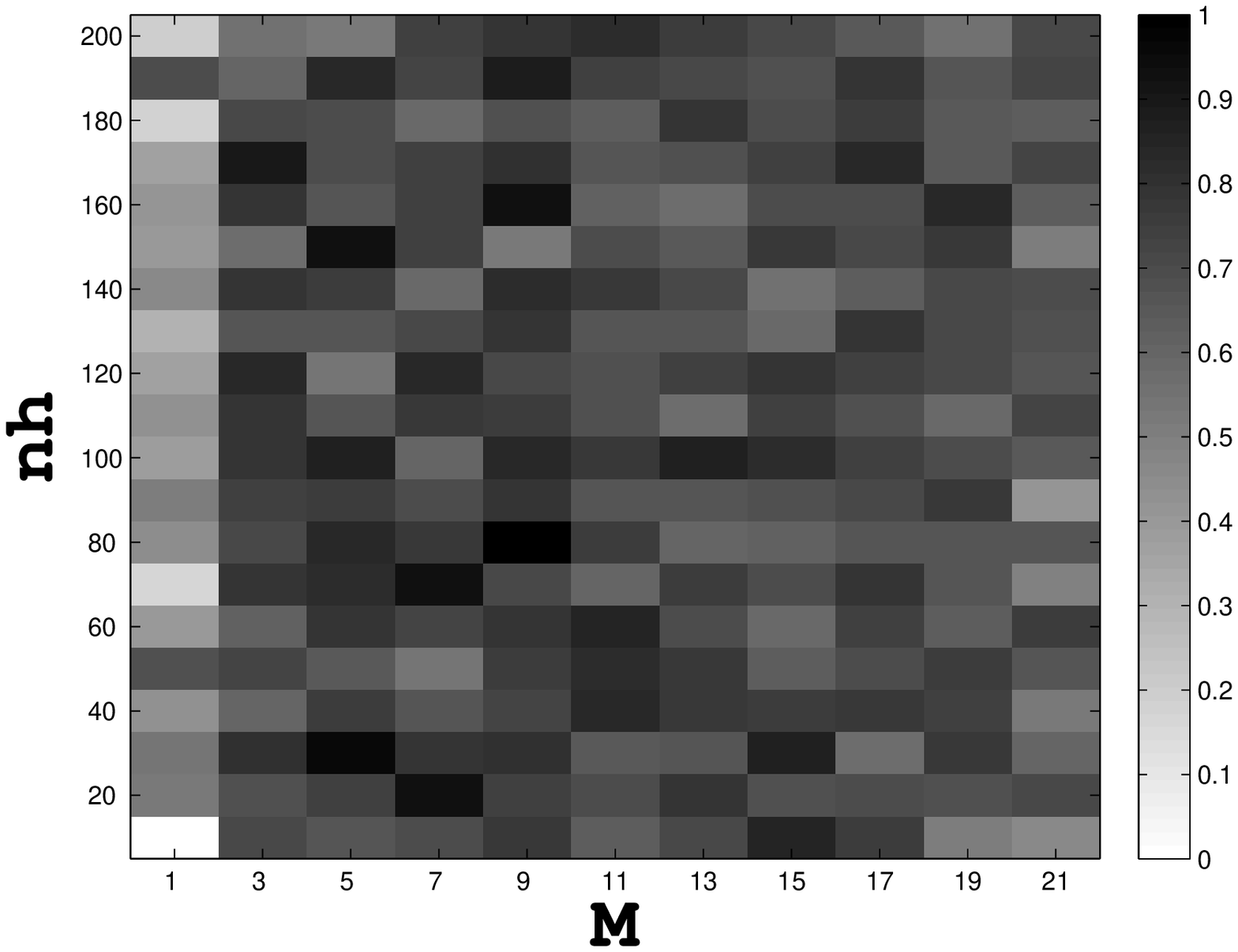}
		\caption{Bölümleme ve gizli düğüm sayısı.}
		\label{fig:pageblocks_map_nh}
	\end{subfigure}
	\hspace{1pt}
	\begin{subfigure}[b]{0.15\textwidth}
	\includegraphics[width=1\linewidth]{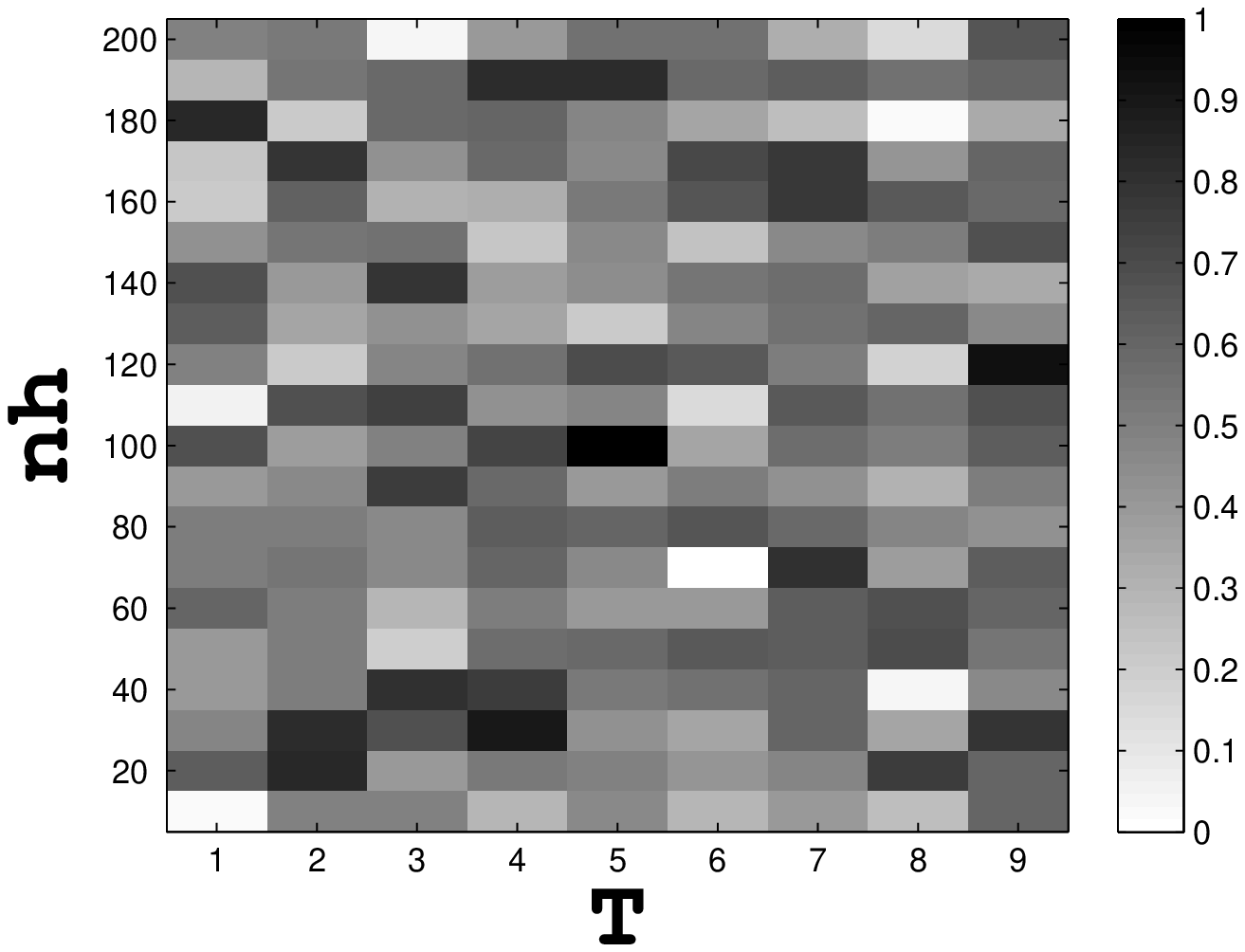}
	\caption{AdaBoost $T$ boyu ve gizli düğüm sayısı.}
	\label{fig:pageblocks_T_nh}
	\end{subfigure}
	\caption{Page blocks veri kümesi ısı haritası.}
	\label{fig:pageblocksres}
\end{figure}

Yapılan ölçümlerin sonuçları Tablo \ref{tbl:bestres}'de gösterilmiştir. Her bir veri seti için bölümleme, $M$, saklı düğüm sayısı, $nh$, AdaBoost iterasyon sayısı, $T$, parametrelerine göre doğruluk değişiminin sonuçları Şekil \ref{fig:statlogres} -- \ref{fig:pageblocksres}'de gösterilmiştir. Isı haritalarında renk, siyaha yaklaşması durumunda modelin doğruluğu artmaktadır. Ölçek her bir grafiğin yanında bulunan renk çubuğu ile verilmiştir.

\begin{table}[!t] %[ph]
	\scriptsize
	\caption{Veri setlerinin en iyi performans sonuçları.}
	\label{tbl:bestres}
	\begin{tabular}[t]{|c|r|r|r|r|r|r|r|}
	\hline \textbf{Veri S.} & \textbf{\# C.} & $T$ & \textbf{\# H.N.} & \textbf{Doğ.} & \textbf{H.} & \textbf{G.C.} & \textbf{$F_1$} \\
	\hline
	Pendigit & 20 & 10 & 21 & 0,8256 & 0,8369 & 0,8234 & 0,8301 \\
	Skin & 21 & 5 & 21 & 0,9892 & 0,9773 & 0,9913 & 0,9842 \\
	Statlog & 11 & 2 & 21 & 0,9103 & 0,7486 & 0,5069 & 0,6045 \\
	Page Blocks & 1 & 1 & 340 & 0,9404 & 0,9027 & 0,5756 & 0,7030 \\
	% Waveform & 19 & 6 & 40 & 0,862 & 0,8680 & 0,8605 & 0,8642 \\
	\hline
	\end{tabular}
\end{table}
\vspace{-5pt}
\vspace{-15pt}
\vspace{-15pt}

\vspace{-10pt}
\section{SONUÇ}
\vspace{-5pt}
Bu çalışmada, MapReduce temelli AdaBoost AÖM algoritması uygulaması, yüksek boyutlu veri setlerinin eğitilmesi için önerilmiştir. Girdi matrisini parçalara ayırarak, önerilen yöntem, AÖM sınıflandırmasının eğitim aşamasının karmaşıklığını azaltmaktadır. Matrisin parçalanması ile yaşanacak olan sınıflandırma performans azalmasının üstesinden gelmek için AdaBoost yöntemi ile desteklenmiştir. Deneysel çalışmalarla elde edilen sonuçlarla, sadece yüksek boyutlu veri setlerinin eğitim karmaşıklığı azalmamakta ayrıca geleneksel AÖM algoritmasının sınıflandırma performansına göre artış yaşanmaktadır.
\vspace{-5pt}

Bu çalışma kapsamında önerilen dağıtık AÖM yöntemi veri parça bölümleme uzunluğu, $M$, AdaBoost yöntemi iterasyon sayısı, $T$, gizli katman düğüm sayısı, $nh$, şeklinde üç parametreye sahiptir. Isı haritası grafiklerinde gösterilen sonuçlara bakarak $M$ ve $T$'nin, $nh$ parametresine göre sınıflandırma performans ölçümüne olan etkisi daha fazla olduğu gözlemlenmektedir.
\vspace{-5pt}

Önerilen yöntem, yüksek boyutlu veri setlerinin karmaşıklığını, matrisi alt parçalara ayırarak, AÖM eğitim aşamasının zorluğunu azaltmaktadır. Tablo \ref{tbl:convelm} ve Tablo \ref{tbl:bestres} karşılaştırıldığında, model karmaşıklık göstergesi olarak düşünülen, $nh$ sayısında azalma olduğu görülmektedir. Bu nedenle, yöntem sadece girdi matrisi karmaşıklığını değil aynı zamanda model karmaşıklığınında azalmasını sağlamaktadır.

%\section*{TEŞEKKÜR}
%Yazarın teşekkür etmek istediği kurum yada kişiler burada belirtilecek.



%\bibliographystyle{IEEEtran}

\end{document}